%% file: main_arxiv.tex
\documentclass[11pt,letterpaper]{article}

\usepackage{lineno,hyperref}

\usepackage{authblk}
\usepackage{helvet}
\usepackage{courier}
\usepackage{tikz}
\usepackage{hyperref}
\usepackage{tabularx}
\usepackage{pgfplots}
\usepackage{amsmath}
\usepackage{lipsum}
\usepackage{soul}
\usepackage{xcolor}
\newcommand\ch[1]{%
  \bgroup
  #1%
  \egroup
}

\usepackage{multirow}
\usepackage{booktabs,rotating}

\providecommand{\keywords}[1]{\textbf{\textit{Keywords --}} #1}
\newcommand{\ignore}[1]{}
\newcommand{\nop}[1]{}
\newcommand*{\eg}{{\em e.g.}}

\newcommand*{\ie}{{\em i.e.}}
\newcommand*{\cf}{{\em c.f.}}

\tikzstyle{textnode} = [rectangle, inner sep=0pt,outer sep=0,execute at begin node={\strut}, font=\small]  
\tikzstyle{hbox} = [rectangle, rounded corners, fill=yellow!70, draw=blue!60]

\AtBeginDocument{%
  \providecommand\BibTeX{{%
    \normalfont B\kern-0.5em{\scshape i\kern-0.25em b}\kern-0.8em\TeX}}}

\title{Reddit Entity Linking Dataset}

\author[1]{Nicholas Botzer}
\author[1]{Yifan Ding}
\author[1,*]{Tim Weninger}

\affil[1]{University of Notre Dame, Notre Dame, IN}
\affil[*]{Corresponding author: Tim Weninger, tweninger@nd.edu}

\begin{document}


\date{}
\maketitle

\begin{abstract}
We introduce and make publicly available an entity linking dataset from Reddit that contains 17,316 linked entities, each annotated by three human annotators and then grouped into Gold, Silver, and Bronze to indicate inter-annotator agreement. We analyze the different errors and disagreements made by annotators and suggest three types of corrections to the raw data. Finally, we tested existing entity linking models that are trained and tuned on text from non-social media datasets. We find that, although these existing entity linking models perform very well on their original datasets, they perform poorly on this social media dataset. We also show that the majority of these errors can be attributed to poor performance on the mention detection subtask. These results indicate the need for better entity linking models that can be applied to the enormous amount of social media text.

\keywords{entity linking, dataset, natural language processing}

\end{abstract}

\maketitle

\section{Introduction}
Entity Linking is the problem of mapping free text with the appropriate entities in a structured knowledge base (KB), such as Wikipedia. Linking natural language text to large knowledge graphs enables applications to make use of rich semantic relationships that may be implied in the natural language, but are explicitly expressed in the knowledge graph. Entity linking is therefore a core task in natural language processing with applications in recommender systems~\cite{de2015semantics}, chatbots~\cite{ghazvininejad2018knowledge}, relation extraction~\cite{ren2017constructing}, and knowledge base completion~\cite{dredze2010entity,shi2018open} among many others.

The entity linking task is generally broken down into subtasks: (1) mention detection and (2) entity disambiguation. Similar to named entity recognition (NER), the mention detection subtask finds substrings from text that represent some entity. Mention detection is itself a wide area of research encompassing semantic role labelling~\cite{strubell2018linguistically}, anaphora resolution~\cite{aktas2018anaphora}, and others in order to consider the various ways that humans express entities in natural language. The entity disambiguation task takes the word or phrase identified from the mention detection task and then identifies an entry in a knowledge base that matches the entity. 

\begin{figure}[t]
    \centering
    \input{figure/arxiv-fig-1}
    \caption{Example entity linking task. Highlights indicate three entity mentions from the surface text, blue boxes denote the linked entity corresponding to each mention.}
    \label{fig:examplefig}
    \vspace{.3cm}
\end{figure}
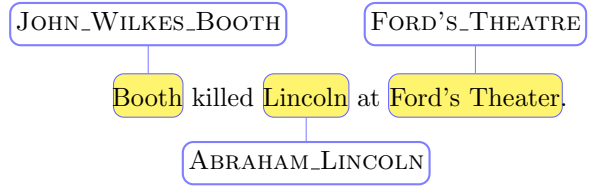

For example, the illustration in Fig.~\ref{fig:examplefig}  shows the phrase ``Booth killed Lincoln at Ford's Theater.'', which has three entity mentions highlighted in yellow: `Booth', `Lincoln', and `Ford's Theater'. These mentions alone are useful for many downstream tasks, but entity disambiguation goes one step further and reconciles the entities with their specific Wikipedia entries (or some other knowledge base) such that the surface text is linked to \textsc{John\_Wilkes\_Booth}, \textsc{Abraham\_Lincoln}, and \textsc{Ford's\_Theatre} respectively. Having linked the surface text with their entities, any number of additional steps can be taken to refine or reason about the data~\cite{chen2019graphflow}.

Because of its wide applicability, entity linking has been heavily explored, and many datasets are available for model training and testing. However, most of these entity linking datasets are extracted from well written news articles, websites, or otherwise professionally curated corpora. Although useful for many tasks, the lexical and linguistic styles of commonly available datasets do not match what is commonly expressed on social media platforms. As a result, recent work has found that many entity linking models do not perform well on social media text~\cite{derczynski2015analysis}.

Besides the lack of training data, entity linking from social media is particularly challenging because of the presence of jargon, poor grammar usage, inconsistent lexical formatting and variation, the wide usage of anaphora, and frequent use of idioms, euphemisms, and other creative (or lazy) use of language. In addition, the use of threading systems in many social discourse platforms adds another layer of complexity into natural language processing systems. 

Some efforts have been made to reconcile this disparity. Typically these systems perform either the mention detection task~\cite{yadav2018survey} \textit{or} the entity disambiguation task~\cite{ganea2017deep, ranAttentionFactorGraph2018, le_improving_2018, leBoostingEntityLinking2019a}, but rarely both. More recently, the rise of deep neural architectures has enabled end-to-end models that learn both tasks jointly~\cite{shimaokaNeuralArchitecturesFinegrained2017,kolitsas2018end}. When considering social media data, most related work use Ritter et al's entity linking dataset from Twitter~\cite{ritter2011named}, which contains hand annotated entity matches for 2,400 Tweets as well as the non-social media datasets \textsc{Figer}~\cite{ling2012fine} and OntoNotes~\cite{gillick2014context} from NIST TAC's entity discovery and linking tasks. Additional hand annotated datasets have been collected from Twitter using various methodologies~\cite{dredze2016twitter,meij2012adding}. 

\begin{figure}
    \centering
    \includegraphics[width=\textwidth]{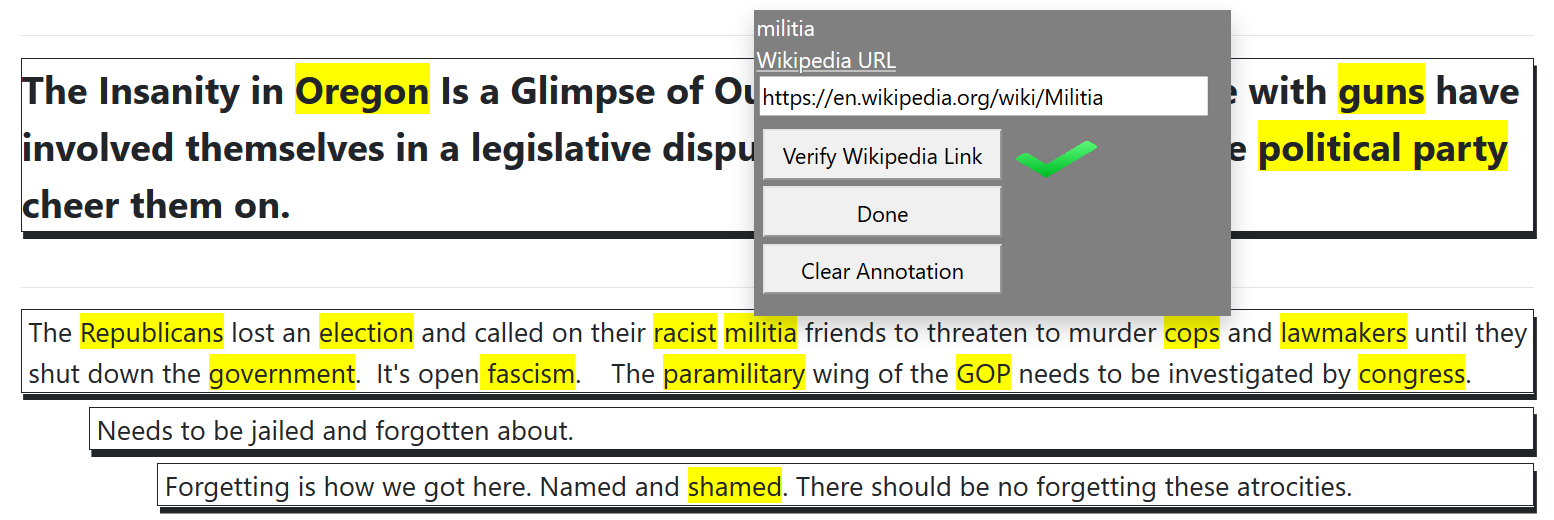}
    \caption{The annotation interface used by human annotators. The interface allowed them to highlight arbitrary spans of text and link them with valid Wikipedia pages.}
    
    \label{fig:application_annotation}
    \vspace{.5cm}
\end{figure}

Although these datasets and algorithms have greatly progressed entity linking, it is important to admit that most social and digital text is not created in Tweet-form. Rather, the vast amount of online communication appears in modern bulletin boards, comment threads, and other user-generated discussion formats. 





In the present work we consider the social media site Reddit. 
Reddit is an archetypal example of a socio-technical discussion board with hundreds of millions of posts and billions of individual comments. On Reddit each post is submitted to a specific subreddit and a discussion is attached to each post. User-provided upvotes and downvotes govern the visibility and ranking of posts and comments on the site. Comments are threaded, meaning that each comment must have a single parent-comment; top-level (\ie, root) comments are considered children of the post text and link, which usually sets the topic of discussion.

Reddit consists of thousands of distinct communities called subreddits.
Each subreddit is associated with a specific topic, such as r/news or r/pokemon, and a volunteer team of moderators that create and enforce guidelines for each subreddit.
Posts made within each subreddit must adhere to these guidelines or offending users may be banned from future posting or commenting in the subreddit. Due to the more relaxed moderation rules on the platform, especially surrounding controversial topics, Reddit threads commonly contain toxic, harassing, or explicit commentary~\cite{mittos2020analyzing}.
A distinct difference that separates Reddit from Twitter or Facebook is how communities form.
On Reddit, users seek out and find subreddits that interest them, whereas on Twitter, Facebook and other social {\em networks}, users build friendship networks and then share posts within this network~\cite{choi2020social}.

Because of its breadth and wide availability, researchers have begun to use Reddit and similar socio-technical discussion boards like HackerNews, Slack, Discord, Disqus for a variety of tasks.
The tasks generally tackle specific subreddits and the community dynamics that arise within them. A significant portion of this research has analyzed subreddits and user behaviors surrounding mental health disorders, which has taken the approach of extracting text and performing clustering to find different themes within these subreddits \cite{park2018examining,yoo2019semantic}. 
In another study, the subreddit r/SuicideWatch was analyzed to help determine whether a user may be at risk of committing suicide \cite{zirikly2019clpsych}.
The resulting data from this allowed researchers to create models to identify users that are at risk in the future.

\ch{Reddit has also proven to be a good source for the collection of datasets. One work in particular utilizes posts from ten subreddits and then has them annotated using Mechanical Turk to determine whether the poster's behavior exhibits stress or not~\cite{turcan_2019_DreadditRedditDataset}. In some cases, the representative labels for a task are built right into Reddit. For example, the common occurrence of sarcasm labels on Reddit can be used to create a large corpus for sarcasm classification~\cite{khodak_2018_LargeSelfAnnotatedCorpus}.}

Furthermore the threaded nature of Reddit posts has allowed researchers to explore how persuasive arguments are carried out~\cite{dutta2020changing}. By analyzing the arguments of the r/Changemyview subreddit researchers were able to capture the argumentative sentences and identify important persuasive discourse markers. 

Generally speaking, social media analysts are beginning to develop and employ text analysis tools to understand a wide variety of social and interpersonal issues such as gender differences \cite{thelwall2019she}, weight loss trends \cite{enes2018reddit}, and the spread of technological innovations \cite{glenski2019characterizing}.


However, the lexical and linguistic style used in online discussion boards is not only different from professional journalistic styles used to curate large natural language processing models and datasets. Some datasets have recently become available for microblogs like Twitter~\cite{manikondaTwitterSparkingMovement2018, priyaWhereShouldOne2019}, but style of microblogs are vastly different from discussion boards. Despite the large amount of social data contributed to these platforms there exists no large, labeled dataset from these types of threaded discussion boards.

\subsection{Related Works}

Within the scope of natural language processing, the problem of entity linking has been quite extensively studied and a vast amount of papers exist discussing the subject; Shen et al. \cite{shen2014entity} and Sevgili et al. \cite{sevgili2020neural} present good surveys of the field. \ch{We also highlight two other works that have previously performed crowd sourced entity linking, \cite{derczynski2015analysis, ide_crowdsourcing_2017} and compare our method against theirs in Section~\ref{sec:experiments}.}  One of the issues for researchers working in this domain is the lack of training and testing data for social media, as well as focus on only narrow portions of the overall problem, \eg, news articles and Wikipedia. In the present work, we review some of the recent entity linking techniques and recommend that researchers focus future work on social discussion boards like Reddit and produce holistic end-to-end models rather than piecemeal approaches to entity linking.

Existing models have major problems with not being able to label entities if they were not included in the initial training set. The concept of ``zero-shot'' learning was developed  to help solve this problem~\cite{logeswaran_zero-shot_2019,shi2018open}. One major benefit of zero-shot learning is that the model can correctly map entities without having trained on any mentions for it.
Others have sought to exploit the type systems of Wikipedia to improve the disambiguation of models by understanding the different contexts for a given mention \cite{onoe2020fine, raiman2018deeptype}. By using the categories of Wikipedia as entity-types, the models can disambiguate the word correctly by using the context of surrounding words.
Another successful approach uses reinforcement learning and a global sequence perspective to select entities for disambiguation \cite{fang2019joint, yang2019learning}. These models are able to move beyond the local context of the type models and use all of the previously predicted entities to improve performance.

Previous works have attempted to tackle the problem on Twitter. Early work on social entity linking has focused on the entity disambiguation task for tweets by utilizing a graph based framework that captures user's topics of interest~\cite{shen2013linking}. Other models have been developed to perform end-to-end entity linking on Twitter. Guo et al. \cite{guo2013link} approach the problem by using a structural SVM that jointly optimizes the mention detection and entity disambiguation task together. Similarly, Fang and Chang \cite{fang2014entity} create an end-to-end spatiotemporal model to extract information from tweets. Unfortunately, the data used in these three papers is not made publicly available.

Finally, researchers have been able to employ deep language models, like BERT, for entity linking~\cite{devlin2019bert}. In one study, BERT was able to achieve good performance on the end-to-end entity linking task with only small changes \cite{broscheit2019investigating}. In another study, Yamada et al. \cite{yamada2020global} trained BERT on a new masked entity prediction task. By using a new pre-training task they are able to embed contextualized entities within BERT allowing for superior performance on a variety of entity based tasks.

Yet, these benchmark datasets are almost entirely based on journalistic quality news text, which are vastly different than the enormous amount of text that is posted to social bulletin boards.

\subsection{Research Questions}

To help alleviate the deficit of training data from social bulletin boards, we introduce and make available a dataset of 17,316 hand-annotated entities from 619 Reddit posts and a sample of comments from each of their comment threads. For each post we asked three annotators from Amazon's Mechanical Turk to: (1) hand-label any and all entity mentions and (2) link each mention to its representative entry on Wikipedia.

Specifically, we ask the following research questions: 

\begin{enumerate}
    \item[RQ1]: How well do human annotators agree on entity labels?
    \item[RQ2]: Do existing state of the art entity linking models perform well on the social discussion dataset?
    \item[RQ3]: Which parts of end-to-end entity linking models are most responsible for errors in entity linking on social text?
\end{enumerate}

We evaluate the agreement rate of different entities and annotators and make simple corrections within the raw data. Next, we evaluate several popular entity linking models and algorithms on this new dataset. Despite our various attempts, we observed that existing models did not outperform simple baselines on this new data. These results indicate that much work remains in the entity linking task. It is our goal that this dataset spurs further development of entity linking systems especially on the vast amount of non-Tweet styled social discussion data.



\begin{table}[t]
    \centering
    \caption{Subreddits from which post and comments were selected for annotation.}
        \vspace{0.1cm}
        \input{tables/new_table-1}
    \label{tab:dataset}
    \vspace{-.2cm}
\end{table}

\section{Collection Methodology}
The dataset introduced in this paper was collected from a variety of posts from various subreddits. Table~\ref{tab:dataset} shows the subreddits from which posts were collected as well as the number of comments. The subreddits were selected based on two criteria, the first being the popularity of the subreddit. The second criteria was subreddits that were likely to include a broad variety of entities.   For each post we selected the top-scoring comment (upvotes minus downvotes) and, when available, the top-scoring child and grand-child comment. Due to the nature of choosing popular subreddits and using high scoring posts our dataset has an emphasize on high quality content selected for annotation. Collection was limited to only those posts submitted to Reddit between Jan. 2018 and Aug. 2019. 


On average, the mean length of a post in our dataset is 20 tokens and the mean length of a comment is 34 tokens. The shortest post was 4 tokens and the longest was 66 tokens. The shortest comment was one token  with the longest comment being 207 tokens. Because we are providing this data to human annotators, we also removed a small number of posts that had explicit content. In total we collected 619 posts and 1,243 comments from 9 subreddits.

\subsection{Human Annotation}
To collect entity mentions and their matches from Reddit posts and comments, we created a straightforward web annotation framework and provided it as a human intelligence task (HIT) to annotators from Mechanical Turk. We limited the HIT to include only American annotators because the posts and comments were English and primarily focused on American current events and culture.  This task was reviewed and approved by the University of Notre Dame's Internal Review Board (\#20-01-5751).
Only after informed consent confirmation was obtained, the annotator was given the following instructions: 

\begin{table}[h!]
    \centering
    {\renewcommand{\arraystretch}{1.3}
    \begin{tabular}{p{0.01\linewidth}p{0.89\linewidth}}
        \hline
    1. & Highlight each portion of text that you believe matches with a Wikipedia article and left click to open a text box. \\
    2. & Determine the correct Wikipedia page for the text based on the context of the discussion thread and copy the Wikipedia link into the Wikipedia URL section. Verify the link entered is a valid Wikipedia URL by clicking the ``Verify Wikipedia Link'' button. \\

    \end{tabular}
    }
\vspace{-0.2cm}
\end{table}

\begin{table}[h!]
    \centering
    {\renewcommand{\arraystretch}{1.3}
    \begin{tabular}{p{0.01\linewidth}p{0.89\linewidth}}
    3. & A green checkmark will appear to the right if the URL is valid. You may close the box with the ``Done'' button after. \\    
    4. & Once you have highlighted and linked all pieces of text that you feel match with a Wikipedia article please advance to the next page.\\
        \hline
    \end{tabular}
    }
\vspace{-0.2cm}
\end{table}

Next, the annotation system asked the annotator to complete a short (and obvious) practice case. Any incorrect or missing annotations were provided to the annotator as feedback. Because the instructions did not explicitly or rigorously define ``entity'', individual interpretations of what is and is not an entity is an important element in our dataset. It has been shown in prior work that the consensus of what constitutes an entity is not concrete, even for researchers within the field \cite{rosales-mendezFineGrainedEvaluationEntity2019}. However, the practice task certainly framed our goal. The practice post was ``\hl{Iran} fires \hl{missile}. \hl{United States} \hl{airbase} struck.'', where the four highlighted entities here indicate the expected entity matches.

After the practice task was complete the annotators were asked to annotate 10 posts and their related comments one at a time. A screen capture of the annotation system is illustrated in Fig.~\ref{fig:application_annotation}. To make an annotation, an annotator selected a substring of text from the post title or comment. Immediately after text is selected a popup-style box would appear with the highlighted text and a prompt to enter the Wikipedia link. Presumably, the annotator would search Wikipedia for a proper entity link and provide it to the system. To ensure consistency, the link had to be verified to resolve with a valid HTTP response from Wikipedia before the annotation could be submitted. 

Each post and comment thread was annotated by three different annotators \ch{as was done in two other crowd sourcing studies \cite{derczynski2015analysis, ide_crowdsourcing_2017}}. By asking the same user to detect entity mentions and perform the linking we expect annotations to vary among the annotators. This task is different from other \ch{crowd sourcing} entity linking methodologies, which pre-selected the entity mentions \cite{derczynski2015analysis} or only asked for entity-links from mentions previously found by other users \cite{ide_crowdsourcing_2017}. \ch{Our methodology also differs in that we use no expert annotators to adjudicate disagreements in the annotations.}

Each annotator was paid \$4.00 USD for their effort. Upon completion we briefly inspected each annotators' submissions to ensure that a reasonable effort was made. Overall we received 17,316 annotations from 202 annotators not including the submissions of 28 annotators that were rejected. In all but the most egregious cases, rejected annotators were still paid. In a small number of instances, annotators did not complete the full set of 10 annotations. We have records of their annotations, but because we cannot verify the quality of these incomplete tasks, we do not include them in our analysis or the released dataset. In total we spent \$908.00 USD for 227 annotators.

\nop{
To ensure that our annotations were of good quality we display each thread grouping to three different annotators as was done by \cite{derczynski_analysis_2015}.
By doing this it allows us to break our annotations down into three groups based on the inter-annotator agreement.
The categories we have broken down the inter-annotator agreement into are gold, silver, and bronze.
Each group denotes a different level of inter-annotator agreement and usage for task as defined in the following sections.
A breakdown of the number entities for each grouping can be seen in table \ref{tbl:dataset_breakdown}.

To perform the annotation we built a web application and used Amazon Mechanical Turk to hire workers.
Our interface instructed users to highlight text and link it with the correct entity on Wikipedia.
We gave a loose definition of the task to annotators but also provided them with an example.
Our definition instructed users to highlight any text that they felt mapped with a Wikipedia page.
As highlighted by other studies on entity linking  \cite{Rosales-Méndez_2019_evaluation} the definition of what is a correct mapping can vary based on the task definition.
\textcolor{red}{By allowing for a loose interpretation our dataset shows what most people interpreted the task and mappings to be. (Fix this section above)}
}

\begin{table}[t]
\centering
\caption{Number of annotated entities from posts and comments. Inter-annotator agreement produces groups of Gold, Silver, and Bronze annotation levels.}
\vspace{0.1cm}
{\renewcommand{\arraystretch}{1}
    \input{tables/table-2}
}
\vspace{.3cm}
\label{tbl:dataset_breakdown}
\end{table}

\subsection{Annotation Results}
The collected results were aggregated to determine the agreement among the three annotators. Gold annotations are those entities identified and linked by all three annotators. For now, Gold annotations must agree on the exact text selection (including punctuation, whitespace, etc) and link to the same Wikipedia page (without considering redirection-pages, disambiguation-pages, etc). Because gold annotations demonstrate unanimous agreement, we consider them to be high quality.

Similarly, silver annotations are those entities selected by two of the three annotators. Although these annotations do not show unanimous agreement, they generally remain high quality and are useful for evaluation and training purposes.

Finally, bronze annotations are those entities indicated by a single annotator. Due to the lack of inter-annotator agreement, bronze annotations are the lowest quality, but still contain interesting occurrences. For example, in the post title beginning with ``Box Office Week: \hl{Black Panther} smashes at \#1 with \$201M...'', all three annotators agreed that the text ``Black Panther'' was an entity mention, but they linked the mention to three different Wikipedia pages: the film, the comic book, and the animal.

Raw results of annotation for both posts and comments are displayed in Table~\ref{tbl:dataset_breakdown}. Note that the number of gold entities is not exactly the same as the number of unique entities and unique mentions. This is because the exact same entity (\ie, Wikipedia page) and mention is sometimes found in two different post titles. Likewise, different surface forms often link to the same entity (\cf, entity resolution), and the same surface form sometimes links to the different entities (\cf, entity disambiguation).

\subsection{Cleaning}
Because it is important to consider how small annotation disagreements and different interpretations permeate this dataset, we performed a critical analysis of annotator disagreement and applied simple procedures to clean certain differences from the dataset.


\subsubsection{Redirection Cleaning} 
Our first cleaning step was to reconcile Wikipedia redirection pages. For example, the entity mention ``Donald Trump'' is frequently linked to the Wikipedia page entitled \textsc{Trump\_(president)}, but this Wiki page automatically redirects to \textsc{Donald\_Trump}. From the annotator's point of view, these two Wiki pages are identical in all ways except for the link. Yet these two different links are considered disagreements in our raw dataset. So, we found all redirection links provided by the annotators and reconciled them to the same Wiki page. This simple cleaning task resulted in an additional 59 gold annotations and an additional 66 silver annotations.

After this simple cleaning, we performed a more-in-depth analysis. We observed two types of errors: link disagreement and mention disagreement.

\subsubsection{Link Disagreement}
Link disagreement occurs when two or more annotators highlight the exact same surface text, but link the text to different entities. We sought to understand this disagreement more thoroughly. Specifically, we asked: When annotators disagree on an entity, are their choices close to one another? That is, do they link to two entities which are similar? Or do their ideas over what the surface form represents diverge significantly?

To answer this question we used the Wikipedia Link Measure (WLM) to measure the similarity between two different entity annotations that have identical surface forms. The WLM score uses the internal link structure of Wikipedia to measure the similarity between two Wiki pages based on how many incoming links the pages share. Formally, given two entity pages $a$ and $b$ as well as their links within the Wikipedia graph $W$, WLM is defined as:

$$
    \textrm{WLM}(a,b) = 1-\frac{\log(\max(|A|,|B|)) - \log(|A \cap B|)} {\log(|W|) - \log(\min(|A|, |B|))}, 
$$

\begin{figure}[t]
    \centering
    \input{figure/wiki_similarity_histogram}
    \caption{Wikipedia Link Measure comparing annotations with Link Disagreements with null model. Despite linking to different pages for the same mention, these Link Disagreements are significantly more similar than random Wikipedia pairs (Mann Whitney $\tau=0.16$, $p<0.001$). }
    \vspace{.4cm}
    \label{fig:wikipedia_similarities}
\end{figure}
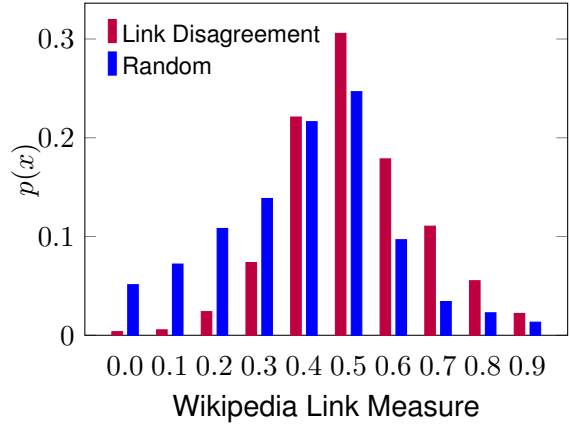

\begin{table*}[t]
\centering
\caption{Top 10 entities from the gold and silver annotations of the dataset. For each entity a list of the mention texts that they were annotated with are shown.}
\begin{tabularx}{\linewidth}{>{\hsize=.6\hsize}X l >{\hsize=1.4\hsize}X}
\toprule 
\textbf{Entity} & \textbf{\#} & \textbf{Unique Mentions} \\ \midrule
\textsc{Donald\_Trump}       &           88      &    President Trump, Trump, trump, President, trumpian, Donald Trump, Tump, US President Donald Trump, Trump's, 45, his             \\ 
\textsc{United\_States}       &        53         &    U.S., US, American, USA, U.S, America's, United States, America             \\ 
\textsc{Film}       &         36        &      Films, films, Movie, movie, FILMS, mocie, movies, Movies, Film, film           \\ 
\textsc{China}       &         30        &      china, Chinese, China           \\ 
\textsc{Russia}       &          26       &      Russia, Russian, Russians           \\ 
\textsc{Human}       &       18          &     human race, Humans, human being, human, Humanity, humans            \\
\textsc{Republican\_Party\_ (United\_States)}      &       16          & Republicans, GOP, red, Republican Party, Republican                \\
\textsc{Germany} & 15 & Germany, germany \\ 
\textsc{Research}       &         14        &   Researchers, research, researcher, Research, researchers              \\ 
\textsc{Scientist} & 14 & scientists, Scientists, scientist \\ 
\bottomrule
\end{tabularx}
\vspace{.5cm}
\label{tab:top_entities_names}
\end{table*}

\noindent where $A$ and $B$ are the sets of incoming links to $a$ and $b$ respectively~\cite{witten2008effective}. Simply put, WLM uses the Wikipedia network to measure the semantic similarity of two entities. We compare the mismatched annotations against a null model consisting of random pairings of entities with the same number of incoming links. Figure~\ref{fig:wikipedia_similarities} illustrates the WLM distribution of mismatched entities against the null model. A Mann-Whitney Test of two distributions determined that the mismatched annotations are significantly more similar than the null model ($\rho=0.16$, $p<0.001$). These results indicate that, although annotators may not explicitly agree on the linked entity, their annotations are much more similar than if they chose a random entity. 

Although we observe that these disagreements link to similar Wikipedia entities, we have no way of reconciling these differences. Therefore we cannot apply any changes to the raw data that correct link disagreements.

\begin{table}[t]
\centering
\caption{Redirection correction (Redir.), mention correction (Men.), and both corrections (All) made to the raw annotation data increase the annotator agreement.}
{\renewcommand{\arraystretch}{1.1}
\begin{tabular}{@{}l l  r  r r r @{}} 
 \toprule
 & &  \textbf{Raw} &   \textbf{Redir.} & \textbf{Men.} & \textbf{All} \\ \midrule
\parbox[t]{2mm}{\multirow{3}{*}{\rotatebox[origin=c]{90}{\textbf{Posts}}}} & \textbf{Gold} & 662 & 696  & 670 & 704 \\
& \textbf{Silver} & 1,152 & 1,154   & 1,159 & 1,159 \\
& \textbf{Bronze} & 2,699 & 2,594   & 2,661  & 2,557 \\
\midrule
\parbox[t]{2mm}{\multirow{3}{*}{\rotatebox[origin=c]{90}{\textbf{Comnts}}}} &\textbf{Gold} & 591 & 616    & 613 & 638 \\
& \textbf{Silver} & 1,544  & 1,552   & 1,559  & 1,564 \\
& \textbf{Bronze} & 4,662 & 4,573  & 4,566  & 4,481 \\
\bottomrule
\end{tabular}
}
\vspace{.1cm}
\label{tab:dataset_corrected}
\end{table}

\subsubsection{Mention Disagreement}
Mention Disagreement occurs when two or more annotators agree on the entity link but disagree on the surface form of the entity mention. Usually, these disagreements differ by only a character or two. In other instances, mentions disagree on whether to include the plural or possessive endings of an entity name. We reconcile these differences by expanding all annotations to end at word boundaries, \eg, white space or punctuation.

Redirection cleaning and mention disagreement correction were applied to each set of annotations. The results displayed in Table~\ref{tab:dataset_corrected} show how the numbers for each annotation level change as Redirection Cleaning (Redir.), and Mention Disagreement correction (Men.) were performed individually. The final column (All) displays the final results with all corrections applied collectively.

\subsubsection{Other Disagreements}

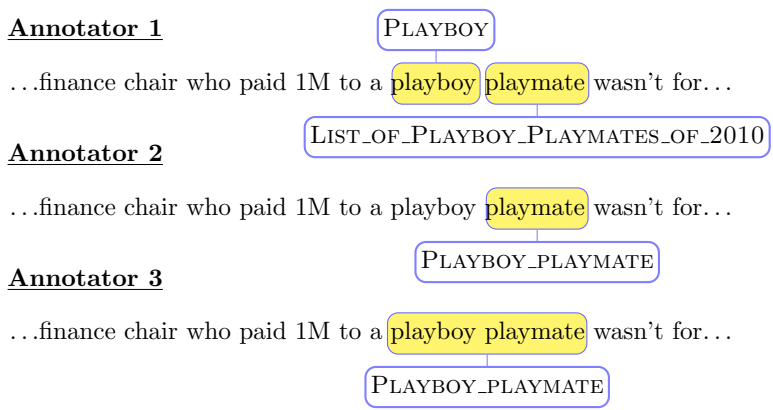
\begin{figure}[t]
\vspace{.4cm}
    \centering
    \input{figure/arxiv-fig-4}
    \caption{Instance where three annotators provided different mention-entity annotations. Annotations, in this case, are extremely close but are not in agreement. Highlights indicate entity mention, blue boxes denote the linked entity from the example text}
    \label{fig:playboy_example}
\end{figure}

Another interesting aspect of this dataset are the details of how annotators selected certain mentions and entities. Consider, for example, an actual disagreement among three annotators illustrated in Fig.~\ref{fig:playboy_example}. In this instance, the four bronze annotations (two from annotator 1, and one each from annotators 2 and 3) show a mix of link disagreement and mention disagreement. Despite their similarity, none of these annotations can be reconciled using the above corrections in part or in aggregate. In annotator 1's case there is no reference to the year 2010 in the text, showing that some annotators may have selected an entity that is somewhat similar. This example shows how difficult finding annotation agreement among human annotators can be. Despite this difficulty, this dataset shows a surprising amount of agreement, and might prove useful to better understanding annotator perception and biases.

\begin{table}[]

\centering
\caption{Results on the five other non-social media datasets for entity disambiguation (ED) and End-to-End Methods (E2E EL). Results are reported for Precision (P), Recall (R), and F1 score.}
{\renewcommand{\arraystretch}{1.1}
\vspace{.1cm}
\tabcolsep=0.11cm
    \input{tables/other_datasets_table}
}
\label{tab:other_datasets}
\end{table}

\subsubsection{Top Entities and their Mentions}
Another way to view this data, as well as the difficulty of the entity linking task in general, is by considering the different mentions that link to the same entities. The entries in Table~\ref{tab:top_entities_names} show the top 10 most frequent entities linked in our dataset as well as some of the mentions that are found to link to them. Many entities, like \textsc{United\_States}, have multiple variations in how they are mentioned in text, especially in social text. Other entities, like \textsc{Film}, are represented by many synonymous terms, as well as different capitalization, misspellings, and other variations in their surface forms.

Although not the objective of the current work, the entity mentions and their links may be a valuable resource for social discourse analysis and other linguistic studies.

\subsection{Data Availability}

The dataset containing the cleaned annotations presented in Table~\ref{tab:dataset_corrected} is available at \url{https://doi.org/10.5281/zenodo.3970806}.

The dataset contains the data of posts and comments gathered from Reddit. Metadata for each post contains the post id, the subreddit to which it was posted and the post's title. Metadata for each comment contains the post id, comment id, subreddit, the parent id (which is itself a comment or post in the dataset), and the comment text. Post and comment ids can be used to gather additional data from Reddit directly. 

Annotations are stored for post and comments according to their agreement: Gold, Silver, or Bronze for a total of 6 annotation files.

In addition to post or comment ids, annotations contains the mention text, the linked Wikipedia entity (after applying redirection correction), the start and end position of the mention from the mention text, and the corrected mention text.


\section{Experiments}
\label{sec:experiments}
Our second goal is to evaluate the performance of entity linking models on our new dataset. Because the bronze annotations fail to show any annotator agreement, we do not perform an evaluation on this part of the dataset. Instead, we limit our evaluation to the gold and silver annotations, which we consider to be of high quality due to their inter-annotator agreement.

As discussed earlier, the entity linking task can be viewed as two separate consecutive tasks: (1) mention detection and (2) entity disambiguation. The first task is widely recognized as named entity recognition (NER) which is not considered individually in this paper. The second task itself has received much attention in recent years, with the introduction of deep neural network-based entity disambiguation models~\cite{ganea2017deep,ranAttentionFactorGraph2018,le_improving_2018, leBoostingEntityLinking2019a, yamada2016joint,fang2019joint}.
However, only a handful of end-to-end models have been developed to perform both tasks jointly~\cite{kolitsas2018end,van2020rel}. 

Overall, we evaluate six different entity disambiguation models and four end-to-end models.

\subsection{Entity Disambiguation Models}
We formally define the entity disambiguation task as follows. Given a document $D = \{w_1, w_2, ... w_n\}$ containing $n$ tokens, a set of entity mentions from the document $M = \{m_1, ..., m_x\}$, and a knowledge base of entities $E = \{e_1, e_2, ... e_m\}$, the goal of entity disambiguation is to find a mapping $\mu: M \mapsto E$ for each mention $m \in M$ to the appropriate entity $e \in E$ in the knowledge base.

\begin{table}
\centering
\caption{Results on the AIDA-CoNNL datasets for entity disambiguation (ED) and end-to-end entity linking (E2E EL). Results are reported for Precision (P), Recall (R), and F1 score.}
{\renewcommand{\arraystretch}{1.1}
\vspace{.1cm}
    \input{tables/aida_datasets_table}
}
\label{tab:aida_dataset}
\end{table}

\begin{table}
\centering
\caption{Results on our Reddit datasets for entity disambiguation (ED) and end-to-end entity linking (E2E EL). Results are reported for Precision (P), Recall (R), and F1 score.}
\vspace{.1cm}
{\renewcommand{\arraystretch}{1.1}
    \input{tables/reddit_results_table}
}
\vspace{.3cm}
\label{tab:reddit_results}
\end{table}

We used six methods: two high-quality baselines and four popular entity disambiguation models.  First, we evaluated entity disambiguation on existing non-social media datasets from AIDA-CoNLL (AIDA-train, AIDA-A, AIDA-B), MSNBC (MSN), AQUAINT (AQNT), ACE2004 (ACE), WNED-WIKI (WIKI), and ClueWeb (ClueW) datasets; as well as the All-corrected Gold, Silver, and Gold + Silver (Comb.) of the Reddit entity linking datasets after corrections were made (\ie, the All-column from Table~\ref{tab:dataset_corrected}). 

Simply put, given a dataset containing text and annotated mentions, the models try to predict the correct Wikipedia entities for each mention.
The details of each model are as follows:
\begin{itemize}
    \item Query: In this baseline method we use Wikipedia's search API\footnote{https://pypi.org/project/wikipedia/} to query each mention. We used the top ranked result as the predicted Wikipedia entity.
    \item Prior: Another common baseline used in entity disambiguation models is the entity-mention prior dictionary $p(e|m)$ obtained from Wikipedia. For this baseline, we simply pick the entity that occurs most frequently for a for each mention.
    \item deep-ed \cite{ganea2017deep}: Uses a combination of entity embeddings, entity-mention prior dictionary, and a contextual attention mechanism to create the local model. Entity embeddings are obtained by considering word-entity co-occurrence. The final global model considers the disambiguation task as a sequential decision problem solved using loopy belief propagation (LBP) with linear conditional random field (CRF). Following the instructions of the model's authors, we retrained deep-ed by adding the mentions of our Reddit dataset into the training set. 
    \item mulrel-nel \cite{le_improving_2018}: Encodes pairwise relations between mentions as latent variables. It also adds dummy mentions to reduce noisy relations. 
    
    \item wnel \cite{leBoostingEntityLinking2019a}: Previous entity disambiguation models rely heavily on the AIDA-train dataset for training. Here the authors introduce a linker model trained on AIDA-train and another unlabeled dataset with generated pseudo labels.

    

    \item End-2-End \cite{kolitsas2018end}: 
    Similar to deep-ed, the End-2-End model generates an entity embedding by considering word-entity co-occurrence in Wikipedia. It first applies an LSTM to build word embeddings by combining word2vec and character embeddings. Then, a global voting algorithm is used by combining word-entity correlation, entity-entity pairwise correlation and a mention-entity prior dictionary. Although the end-to-end model can be used to perform both the mention detection and entity disambiguation tasks, in these first experiments we use only the entity disambiguation portion of the model.
    
    

\end{itemize}

All of the non-baseline models, mulrel-nel, wnel, deep-ed, End-2-End, were trained on the training set (AI-Trn) of AIDA-CoNLL dataset~\cite{hoffart2011robust}. AIDA-CoNLL also contains validation (AI-A), and test (AI-B) sets.

Following the advice of the authors, we retrained the deep-ed and End-2-End models and expanded the entity sets using the mentions of our dataset. 
If we did not do this then the unseen mentions and entities from our Reddit dataset would not be available to deep-ed and End-2-End to disambiguate. Unfortunately, this entity expansion results in a slight performance decrease on the original, non-social media, datasets.

We measure the micro-precision, micro-recall, and micro-$F_1$ scores on the predicted entities compared to the ground truth. For the ED task, precision is simply the number of correct predictions out of all predictions made. Recall is measured as the number of correct predictions out of the total number of ground truth entities. 
It is important to note that most models will not make a prediction for a mention if the model can not find corresponding entity in the knowledge base. 

We report the results of the models trained non-social datasets in Table~\ref{tab:other_datasets}; the results of the AIDA-CONLL dataset are reported in Table~\ref{tab:aida_dataset}; and results of the various models trained on the new Reddit dataset is reported in Table~\ref{tab:reddit_results}. Results on the Reddit dataset specifically describe the performance of the models on the All-corrected Gold and Silver annotations presented in the current work, as well as the union of these two groups (Comb.).

Two important conclusions can be reached from these results. First, the simple Query baseline model works surprisingly well on the Reddit annotation dataset, but this performance decreases as Silver annotations are included in the dataset. Second, the deep neural network models do not substantially outperform the baseline models on our Reddit dataset. The Query baseline we used achieves the highest $F_1$ score of all methods on our combined dataset -- even higher than the best neural model. An interesting effect can also be seen in the precision of the neural End-2-End model: it achieves the highest precision of all methods but also the lowest recall. This indicates that the model is likely missing many of the entities that are used in our Reddit dataset. In sum, these results show that existing models trained and tuned with typical entity annotations simply do not transfer to Reddit data.

\subsection{End-to-end Entity Linking Models}
Our next goal is to evaluate existing end-to-end models on the overall entity linking task, which is a combination of the mention detection and entity disambiguation tasks. There has been much less work on the end-to-end task compared to the entity disambiguation task, and there are very few systems with available source code. Despite the dearth of models, the end-to-end task is the more realistic scenario because entity mentions are difficult and costly to annotate in text data. In addition, the end to end task is typically more difficult because errors can accumulate through the end-to-end pipeline~\cite{luo2015joint}.

We formally define the end-to-end task as follows. Given a document $D = \langle w_1, w_2, ..., w_n\rangle$ containing a sequence of $n$ tokens, and a knowledge base of entities $E = \{e_1,e_2, ..., e_m\}$. The goal of end-to-end entity linking is to find a list of mention entity pairs $L = {(m_1 , e_1),...,(m_k,e_k)}$ where each mention $m_k = \langle w_a, ..., w_b\rangle \subseteq D$ is correctly mapped to its representative entity $e_k \in E$. 

For the end-to-end task we identified two high-quality baselines and two popular deep neural network models that uses the same dataset from the entity disambiguation task. Given text \textit{without} any annotated mentions each model returns a list of entity mentions and an associated Wikipedia entity. The details of each model are as follows:
\begin{itemize}
    \item NER + Query: In this baseline method we use Stanford's Named Entity Recognition package~\cite{Qi_Zhang_Zhang_Bolton_Manning_2020} to first perform entity mention detection, and then provide each discovered entity to Wikipedia's search API to query each mention. We use the top ranked result as the predicted Wikipedia entity.
    \item NER + Prior: We use Stanford's NER for entity mention detection and then pick the most frequently appearing Wikipedia entity.
    \item End-2-End \cite{kolitsas2018end}: We retrain the end-to-end model by adding Reddit dataset into test datasets.
    \item REL \cite{van2020rel}: We use REL, which combines a state-of-the-art NER method \texttt{flair} \cite{akbik2018contextual}  with the entity disambiguation approach of Le and Titov \cite{le_improving_2018} to create an end-to-end technique.
\end{itemize}

As before, we again measure the micro-precision, micro-recall, and micro-$F_1$ of predicted entities. Following the evaluation methodology used in Kolistas et al. we utilized strong matching, \ie, the mention and entity must match exactly to count as a correct prediction~\cite{kolitsas2018end}. Strong matching also requires that the detected mentions are not allowed to overlap. One important difference between the end-to-end precision and the disambiguation precision is how missed predictions are counted. In the entity linking task, any prediction not in the ground truth counts as a miss, even though it may actually be a correct, albeit missing, prediction. 
Therefore, when we evaluate on the gold and silver Reddit datasets separately they have a lower precision than on the combined dataset. The recall value for end-to-end evaluation is simply the number of correct matches out of the total number of ground truth mention-entity pairs.

We report the results of the end-to-end experiments at the bottom portion of Tables~\ref{tab:other_datasets}, \ref{tab:aida_dataset}, and \ref{tab:reddit_results} for the non-social, AIDA, and Reddit datasets respectively. We observe that existing models, even the deep neural network models, generally exhibit poor performance on the Reddit annotation dataset. We also observe that both of our baseline models perform remarkably worse in precision compared to the neural methods, but the recall measurements are comparable.

We prefer to focus our attention on the recall measurements of these experiments because of the issues with precision previously discussed. However, it is clear that the neural models exhibit fairly high precision and fairly low recall rates on the combined dataset. These results indicate that the advanced neural models are rather conservative in their predictions. Comparing recall rates of social and non-social datasets, it appears that the existing models are unable to find the non-standard and non-popular entities that are a common occurrence in social data.

\begin{table}
\centering
\caption{Analysis of the two different error types, entity errors (E) and mention errors (M), for end-to-end models on the Reddit dataset.}
\vspace{.1cm}
{\renewcommand{\arraystretch}{1.1}
    \input{tables/reddit_errors}
}
\vspace{.3cm}
\label{tab:reddit_errors}
\end{table}

Because of this poor performance of existing models, we were curious to investigate the primary cause of error. In general, we find two possible causes: mention detection error and disambiguation error. Mention detection errors occur when the model fails to detect an appropriate mention from the text. Disambiguation errors occur when a mention is correctly identified, but the wrong entity is linked. We report these two types of errors on the Reddit dataset in Table~\ref{tab:reddit_errors}. These results indicate that the  majority of the errors occur when the model fails to identify a mention. Although this is unsurprising given that these models were trained on traditional entity linking datasets, these results clearly indicate a need for better social media entity linking models.

\section{Discussion and Implications}

To summarize, we return to the three research questions raised in the present work.

First (RQ1), despite the open-ended nature of this task, we found a surprising agreement in the annotations. We showed that human annotators largely agree on entity labels in social media discourse. Human annotations on 619 posts found that all three users agreed on an average of one entity per post title, and two of the three annotators agreed on an additional two entities per post. We also provided some simple corrections to the raw annotations to further boost this agreement. 

We expect that the dataset constructed from this task, which is extracted from hundreds of social media threads and hand-annotated with hundreds of human annotators, will be useful for studying social commentary and discussion. The primary differences between the Reddit Entity Linking dataset and existing datasets are that (i) this Reddit dataset has many updated pop culture references, idioms, and memes that are not present in existings data sources; and (ii) compared to news sources, Wikipedia and other more-professional data sources used to create existing models, this Reddit dataset has significantly more noise, typos, unorthodox spellings, and abbreviations. These unique qualities make this new dataset a possible source for computational social scientists and computational linguists in their development of robust entity linking tools and in their study of social commentary.

Second (RQ2), we show that standard datasets commonly used to to train entity linking models are not able to translate to social media data. The primary problem with the existing models is that they are typically trained on journalistic-quality datasets. However, social media data is far less polished and only sometimes conforms to traditional sentence structure and grammar. In addition, existing datasets do not contain the type of thread-structure that is common in social media discussions. Discussion threads (and the threading system in general) is a relatively new rhetorical mechanism with many undiscovered and poorly understood implications in natural language discourse. It is clear that an entity linking dataset annotated from social media data is needed so that automatic entity linking models can be developed and adapted to this large and growing corpus of human discourse. 

Third (RQ3), by performing an unbiased comparison of various entity linking models, we are able to discern which parts of the models are most responsible for errors in entity linking. Interestingly, we show that baseline models (\ie, simply querying the mention text) performs surprisingly well on the dataset. With that in mind, current works on mention detection and entity disambiguation tasks have been recently combined into end-to-end tasks are mostly based on the prior probability. The utility of a model tends to exist in the addition of local and global context information. However, for social media data, this technique is difficult because priors on social media data requires significant human annotation and can quickly become out of date. 

Comparing the results of the end-2-end model against the NER baselines and entity disambiguation baselines, we find that the majority of the error can be attributed to the mention detection sub-task. That is, identifying which word or words represent an entity mention in social media text appears to be very challenging. But once these mentions are identified, linking them to their proper entity is less challenging. 

Having identified these limitations with the state of the art and armed with the Reddit Entity Linking dataset, the goal of our future work will be to create better end-to-end entity linking models that operate on social media discourse.

\section{Acknowledgements}
We thank Trenton Ford for helping to prepare this manuscript. This work is funded by the US Army Research Office (W911NF-17-1-0448) and the US Defense Advanced Research Projects Agency (DARPA W911NF-17-C-0094).



\end{document}

%% file: figure/arxiv-fig-1.tex
\begin{tikzpicture}[transform shape]
\node [hbox, minimum width=2.4em, minimum height=1.5em] (v2) at (-4.55,0.55) {};  
\node [hbox, minimum width=2.95em, minimum height=1.5em] (v5) at (-2.45,0.55) {};  
\node [hbox, minimum width=6.0em, minimum height=1.5em] (v4) at (-0.20,0.55) {};  
\node [textnode] at (-2,0.5) {Booth killed Lincoln at Ford's Theater.};

\node [textnode, rounded corners, draw=blue!50, thick, inner sep=2pt] (v1) at (-4.55,1.5) {\textsc{John\_Wilkes\_Booth}};
\node [textnode, rounded corners, draw=blue!50, thick, inner sep=2pt] (v3) at (-0.20,1.5) {\textsc{Ford's\_Theatre}};
\node [textnode, rounded corners, draw=blue!50, thick, inner sep=2pt] (v6) at (-2.45,-0.35) {\textsc{Abraham\_Lincoln}};

\draw [draw=blue!50]  (v1) edge (v2);
\draw [draw=blue!50] (v3) edge (v4);
\draw [draw=blue!50] (v5) edge (v6);
\end{tikzpicture}

%% file: tables/new_table-1.tex
\begin{tabular}{@{}lcrr@{}}
    \toprule 
    \textbf{Subreddit} & \phantom{a} & \textbf{Posts} & \textbf{Comments} \\ 
    \midrule
    r/movies     && 67 & 143 \\ 
    r/worldnews     && 158 & 315 \\ 
    r/gaming     && 30 & 63 \\ 
    r/news     && 84 & 163 \\ 
    r/politics     && 71 & 145 \\ 
    r/explainlikeimfive     && 49 & 90 \\ 
    r/sports     && 72 & 148 \\ 
    r/science     && 47 & 91 \\ 
    r/Economics     && 41 & 85 \\ 
    \midrule
    Total && 619 & 1,243 \\
    \bottomrule
\end{tabular}

%% file: tables/table-2.tex
{
\renewcommand{\arraystretch}{1.2}
\begin{tabular}{@{}l l  r r r@{}}
\toprule 
 & \multirow{2}{*}{\textbf{Group}} & \multicolumn{1}{c}{\textbf{Raw}} &   \multicolumn{1}{c}{\textbf{Unique}} & \multicolumn{1}{c}{\textbf{Unique}} \\
&  & {\textbf{Entities}} & {\textbf{Entities}} & {\textbf{Mentions}} \\ \midrule
\parbox[t]{2mm}{\multirow{3}{*}{\rotatebox[origin=c]{90}{\textbf{Posts}}}} & \textbf{Gold} & 704 & 527 & 582 \\
& \textbf{Silver} & 1,159 & 900 & 989 \\
& \textbf{Bronze} & 2,557 & 1,877 & 1,938 \\
\midrule
\parbox[t]{2mm}{\multirow{3}{*}{\rotatebox[origin=c]{90}{\textbf{Comnts}}}} &\textbf{Gold} & 638 & 487 & 527 \\
& \textbf{Silver} & 1,564 & 1,126 & 1,244 \\
& \textbf{Bronze} & 4,481 & 2,788 & 3,3013 \\
\bottomrule
\end{tabular}
}

%

%% file: figure/wiki_similarity_histogram.tex
    \begin{tikzpicture}
    \pgfplotsset{compat=1.11,
    /pgfplots/ybar legend/.style={
    /pgfplots/legend image code/.code={%
       \draw[##1,/tikz/.cd,yshift=-0.25em]
        (0cm,0cm) rectangle (3pt,0.8em);},
   },
}
    
 \begin{axis}
[
    ybar,  
    /pgf/bar width=4pt,  
    width=8cm, 
    height=6cm,   
    symbolic x coords={bin0,bin1,bin2,bin3,bin4,bin5,bin6,bin7,bin8,bin9},
    xticklabels={$0.0$,$0.1$,$0.2$,$0.3$, $0.4$, $0.5$, $0.6$, $0.7$, $0.8$, $0.9$},
    x tick label style={anchor=north},
    xtick style={draw=none},
    xtick=data,
    ymin=0, 
    ylabel=$p(x)$,
    ylabel style={yshift=-0.3cm},
    xlabel=\textsf{Wikipedia Link Measure},
    legend style={draw=none},
    legend pos=north west,
    legend cell align={left}
]

   \addplot+[color=purple] plot coordinates {
   (bin0, 0.003683241252) 
   (bin1, 0.005524861878) 
   (bin2, 0.02394106814) 
   (bin3, 0.07366482505) 
   (bin4, 0.2209944751) 
   (bin5, 0.3057090239) 
   (bin6, 0.1786372007) 
   (bin7, 0.1104972376) 
   (bin8, 0.05524861878) 
   (bin9, 0.02209944751) 
   };
   
   \addplot+[color=blue] plot coordinates{ 
   (bin0, 0.05123339658) 
   (bin1, 0.07210626186) 
   (bin2, 0.1081593928) 
   (bin3, 0.1385199241) 
   (bin4, 0.2163187856) 
   (bin5, 0.2466793169) 
   (bin6, 0.09677419355) 
   (bin7, 0.03415559772) 
   (bin8, 0.02277039848) 
   (bin9, 0.01328273245) 
   };

   \legend{\footnotesize{\textsf{Link Disagreement}}, \footnotesize{\textsf{Random}}}
   
 \end{axis}
\end{tikzpicture}

%
%

%% file: figure/arxiv-fig-4.tex
\begin{tikzpicture}[scale=0.95, transform shape]
	\begin{scope}
		\node [hbox, minimum width=3.2em, minimum height=1.5em] (v2) at (0.91,0) {};  
		\node [hbox, minimum width=3.75em, minimum height=1.5em] (v4) at (2.325,0) {};  
		\node[textnode] at (0, 0) {$\ldots$finance chair who paid 1M to a playboy playmate wasn't for$\ldots$};
		\node [textnode] at (-4,0.75) {\textbf{\underline{Annotator 1}}};
		\node [textnode, rounded corners, draw=blue!50, thick, inner sep=2pt] (v1) at (0.91,0.75) {\textsc{Playboy}};
		\node [textnode, rounded corners, draw=blue!50, thick, inner sep=2pt] (v3) at (2.325,-0.75) {\textsc{List\_of\_Playboy\_Playmates\_of\_2010}};
	
		\draw [draw=blue!50]  (v1) edge (v2);
		\draw [draw=blue!50] (v3) edge (v4);
	\end{scope}
	
	\begin{scope}[shift={(0,1.25)}]
		\node [hbox, minimum width=3.7em, minimum height=1.5em] (v1) at (2.325,-3) {};  
		\node[textnode] at (0,-3) {$\ldots$finance chair who paid 1M to a playboy playmate wasn't for$\ldots$};
		\node [textnode] at (-4,-2.25) {\textbf{\underline{Annotator 2}}};
		\node [textnode, rounded corners, draw=blue!50, thick, inner sep=2pt] (v2) at (2.325,-3.75) {\textsc{Playboy\_playmate}};
		
		\draw [draw=blue!50] (v1) edge (v2);
	\end{scope}
	
	\begin{scope}[shift={(0,2.5)}]
		\node [hbox, minimum width=7.25em, minimum height=1.5em] (v1) at (1.625,-6) {};  
		\node[textnode] at (0,-6) {$\ldots$finance chair who paid 1M to a playboy playmate wasn't for$\ldots$};
		\node [textnode] at (-4,-5.25) {\textbf{\underline{Annotator 3}}};
		
		\node [textnode, rounded corners, draw=blue!50, thick, inner sep=2pt, outer sep=0pt] (v2) at (1.625,-6.75) {\textsc{Playboy\_playmate}};
		
		\draw [draw=blue!50] (v1) edge (v2);
	\end{scope}
\end{tikzpicture}

%% file: tables/other_datasets_table.tex
\small{
\begin{tabular}{ll ccc c ccc c ccc c ccc c ccc}
\toprule

& \multirow{2}{*}{\textbf{Model}} &  \multicolumn{3}{c}{\textbf{MSN}} & &  \multicolumn{3}{c}{\textbf{AQNT}} & & \multicolumn{3}{c}{\textbf{ClueW}} & & \multicolumn{3}{c}{\textbf{WIKI}} & &  \multicolumn{3}{c}{\textbf{ACE}}\\

&  & \textbf{P} & \textbf{R} & \textbf{F$_1$} & & \textbf{P} & \textbf{R} & \textbf{F$_1$} & & \textbf{P} & \textbf{R} & \textbf{F$_1$} & & \textbf{P} & \textbf{R} & \textbf{F$_1$} & & \textbf{P} & \textbf{R} & \textbf{F$_1$} \\
\midrule
\multirow{6}{*}{\rotatebox[]{90}{\textbf{ED}}} & \textbf{Prior} & 0.76 & 0.76 & 0.76 & & 0.86 & 0.83 & 0.84 & & 0.67 & 0.67 & 0.67 & & 0.64 & 0.64 & 0.64 & & 0.90 & 0.84 & 0.87 \\
& \textbf{Query} & 0.64 & 0.64 & 0.64 & & 0.82 & 0.82 & 0.82 & & 0.56 & 0.56 & 0.56 & & 0.57 & 0.57 & 0.57 & & 0.70 & 0.70 & 0.70 \\
& \textbf{deep-ed} & 0.92 & 0.92 & 0.92 & & 0.90 & 0.87 & 0.89 & & 0.76 & 0.76 & 0.76 & & 0.74 & 0.74 & 0.74 & & 0.90 & 0.84 & 0.87 \\
& \textbf{End-to-End} & 0.94 & 0.90 & 0.92 & & 0.92 & 0.87 & 0.90 & & 0.83 & 0.72 & 0.77 & & 0.78 & 0.71 & 0.74 & & 0.93 & 0.84 & 0.88 \\
& \textbf{mulrel-nel} & 0.93 & 0.93 & 0.93 & & 0.88 & 0.85 & 0.87 & & 0.77 & 0.77 & 0.77 & & 0.77 & 0.77 & 0.77 & & 0.91 & 0.85 & 0.88 \\
& \textbf{wnel} & 0.93 & 0.92 & 0.92 & & 0.92 & 0.89 & 0.91 & & 0.77 & 0.77 & 0.77 & & 0.75 & 0.75 & 0.75 & & 0.90 & 0.84 & 0.87 \\
\midrule
\multirow{4}{*}{\rotatebox[]{90}{\textbf{E2E EL}}} & \textbf{NER + Prior} & 0.24 & 0.47 & 0.32 & & 0.16 & 0.34 & 0.22 & & 0.10 & 0.38 & 0.16 & & 0.10 & 0.28 & 0.15 & & 0.08 & 0.84 & 0.14 \\
& \textbf{NER + Query} & 0.24 & 0.47 & 0.32 & & 0.16 & 0.34 & 0.22 & & 0.10 & 0.38 & 0.16 & & 0.10 & 0.28 & 0.15 & & 0.05 & 0.55 & 0.09 \\
& \textbf{End-to-End} & 0.75 & 0.73 & 0.74 & & 0.34 & 0.39 & 0.37 & & 0.44 & 0.48 & 0.46 & & 0.39 & 0.40 & 0.39 & & 0.19 & 0.68 & 0.30 \\
& \textbf{REL} & 0.77 & 0.67 & 0.72 & & 0.37 & 0.29 & 0.33 & & 0.56 & 0.27 & 0.36 & & 0.47 & 0.33 & 0.39 & & 0.58 & 0.17 & 0.26 \\
\bottomrule
\end{tabular}
}

%% file: tables/aida_datasets_table.tex
\begin{tabular}{ll ccc c ccc c ccc}
\toprule
& \multirow{2}{*}{\textbf{Model}} & \multicolumn{3}{c}{\textbf{AIDA-Train}} & & \multicolumn{3}{c}{\textbf{AIDA-A}} & & \multicolumn{3}{c}{\textbf{AIDA-B}} \\
 &  & \textbf{P} & \textbf{R} & \textbf{F$_1$} & & \textbf{P} & \textbf{R} & \textbf{F$_1$} & & \textbf{P} & \textbf{R} & \textbf{F$_1$} \\
 \midrule
\multirow{6}{*}{\rotatebox[]{90}{\textbf{ED}}} & \textbf{Prior} & 0.75 & 0.75 & 0.75 & & 0.71 & 0.71 & 0.71 & & 0.69 & 0.69 & 0.69 \\
& \textbf{Query} & 0.63 & 0.63 & 0.63 & & 0.61 & 0.61 & 0.61 & & 0.60 & 0.60 & 0.60 \\
& \textbf{deep-ed} & 0.87 & 0.87 & 0.87 & & 0.87 & 0.87 & 0.87 & & 0.87 & 0.87 & 0.87 \\
& \textbf{End-to-End} & 0.97 & 0.96 & 0.97 & & 0.95 & 0.93 & 0.94 & & 0.89 & 0.85 & 0.87 \\
& \textbf{mulrel-nel} & 0.95 & 0.95 & 0.95 & & 0.92 & 0.92 & 0.92 & & 0.93 & 0.93 & 0.93 \\
& \textbf{wnel} & 0.86 & 0.86 & 0.86 & & 0.81 & 0.81 & 0.81 & & 0.81 & 0.81 & 0.81 \\
\midrule
\multirow{4}{*}{\rotatebox[]{90}{\textbf{E2E EL}}} & \textbf{NER + Prior} & 0.42 & 0.74 & 0.53 & & 0.41 & 0.71 & 0.52 & & 0.40 & 0.69 & 0.51 \\
& \textbf{NER + Query} & 0.27 & 0.48 & 0.34 & & 0.27 & 0.46 & 0.34 & & 0.25 & 0.44 & 0.33 \\
& \textbf{End-to-End} & 0.96 & 0.96 & 0.96 & & 0.90 & 0.89 & 0.90 & & 0.84 & 0.81 & 0.82 \\
& \textbf{REL} & 0.84 & 0.73 & 0.78 & & 0.80 & 0.71 & 0.75 & & 0.80 & 0.71 & 0.75 \\
\bottomrule
\end{tabular}

%% file: tables/reddit_results_table.tex
\begin{tabular}{ll ccc c ccc c ccc}
\toprule
& \multirow{2}{*}{\textbf{Model}} & \multicolumn{3}{c}{\textbf{Gold}} & & \multicolumn{3}{c}{\textbf{Silver}} & & \multicolumn{3}{c}{\textbf{Comb.}} \\

&  & \textbf{P} & \textbf{R} & \textbf{F$_1$} & & \textbf{P} & \textbf{R} & \textbf{F$_1$} & & \textbf{P} & \textbf{R} & \textbf{F$_1$} \\ \midrule
\multirow{6}{*}{\rotatebox[]{90}{\textbf{ED}}} & \textbf{Prior} &	0.81 &	0.78 &	0.79 & &	0.69 &	0.66 &	0.68 & &	0.73 &	0.70 &	0.72 \\
 & \textbf{Query} & 0.89 & 0.89 & 0.89 & & 0.77 & 0.77 & 0.77 & & 0.81 & 0.81 & 0.81 \\
 & \textbf{deep-ed} & 0.81 & 0.78 & 0.80 & & 0.73 & 0.70 & 0.71 & & 0.76 & 0.73 & 0.74 \\
 & \textbf{End-2-End} &	0.93 & 0.51 & 0.66 & & 0.89 & 0.40 & 0.55 & & 0.90 & 0.47 & 0.62 \\
 & \textbf{mulrel-nel} & 0.72 & 0.70 & 0.71 & & 0.63 & 0.61 & 0.62 & & 0.67 & 0.65 & 0.66 \\
 & \textbf{wnel} & 0.87 & 0.83 & 0.85 & & 0.79 & 0.74 & 0.77 & & 0.83 & 0.78 & 0.80 \\
 \midrule
 \multirow{4}{*}{\rotatebox[]{90}{\textbf{E2E EL}}} 
 & \textbf{NER + Prior} & 0.13 & 0.29 & 0.18 && 0.14 & 0.15 & 0.15 & & 0.28 & 0.19 & 0.23 \\
 & \textbf{NER + Query} & 0.14 & 0.30 & 0.19 & & 0.14 & 0.15 & 0.15 & & 0.28 & 0.22 & 0.24\\
 & \textbf{End-2-End} & 0.22 & 0.21 & 0.22 & & 0.27 & 0.12 & 0.17 && 0.49 & 0.15 & 0.23 \\
 & \textbf{REL} & 0.20 & 0.35 & 0.25 && 0.24 & 0.20 & 0.22 && 0.44 & 0.25 & 0.32 \\
 
\bottomrule
\end{tabular}

%% file: tables/reddit_errors.tex
\begin{tabular}{ll cc c cc c cc}
\toprule
& \multirow{2}{*}{\textbf{Model}} & \multicolumn{2}{c}{\textbf{Gold}} & & \multicolumn{2}{c}{\textbf{Silver}} & & \multicolumn{2}{c}{\textbf{Comb.}} \\
 & & \textbf{E} & \textbf{M} & & \textbf{E} & \textbf{M} & & \textbf{E} & \textbf{M} \\
 \midrule
\multirow{4}{*}{\rotatebox[]{90}{\textbf{E2E EL}}} & \textbf{NER + Prior} & 75 & 881 && 135 & 2179 & & 210 & 3060 \\
 & \textbf{NER + Query} & 61 & 881 && 156 & 2179 & & 217 & 3060 \\
 & \textbf{End-2-End} & 17 & 1048 && 19 & 2369 && 36 & 3418 \\
 & \textbf{REL} & 70 & 803 & & 143 & 2027 & & 213 & 2830 \\
\bottomrule
\end{tabular}